\documentclass[letterpaper, 10 pt, conference]{ieeeconf}  % Comment this line out if you need a4paper

\IEEEoverridecommandlockouts                              % This command is only needed if 
\usepackage{amsmath,amsfonts}
\usepackage{algorithmic}
\usepackage{algorithm}
\usepackage{array}
\usepackage[caption=false,font=normalsize,labelfont=sf,textfont=sf]{subfig}
\usepackage{textcomp}
\usepackage{stfloats}
\usepackage{url}
\usepackage{verbatim}
\usepackage{graphicx}
\usepackage{cite}
\usepackage[normalem]{ulem}
\usepackage{xcolor}
\usepackage{multirow}
\usepackage{booktabs}  
\usepackage{graphicx}
\usepackage{epstopdf}
\usepackage{pifont}

\usepackage{helvet}
\usepackage{threeparttable}

\usepackage{etoolbox}

\title{\LARGE \bf
PIE: \underline{P}erception and \underline{I}nteraction \underline{E}nhanced End-to-End Motion Planning for Autonomous Driving
}

\makeatletter
\patchcmd{\@makecaption}
  {\\}
  {:\ }
  {}
  {}
\makeatother

\author{Chengran Yuan$^{1}$$^*$, Zijian Lu$^{1}$$^*$, Zhanqi Zhang$^{1}$$^*$,  Yimin Zhao$^{2}$, Zefan Huang$^{1}$,  Shuo Sun$^{1}$, \\Jiawei Sun$^{1}$, Jiahui Li$^{1}$,  
    Christina Dao Wen Lee$^{1}$, Dongen Li$^{1}$ and Marcelo H. Ang Jr.$^{1}$
\thanks{$^{*}$ Joint first author}
\thanks{$^{1}$Department of Mechanical Engineering, National University of Singapore, Singapore 119077 (e-mail: \{chengran.yuan\}@u.nus.edu; mpeangh@nus.edu.sg).}% <-this % stops a space
\thanks{$^{2}$Department of Civil and Environmental Engineering, National University of Singapore, Singapore 119077}
}

\begin{document}

\maketitle
\thispagestyle{empty}
\pagestyle{empty}

%%%%%%%%%%%%%%%%%%%%%%%%%%%%%%%%%%%%%%%               %%%%%%%%%%%%%%%%%%%%%%%%%%%%%%%%%%%%%%%%%%%%%

\begin{abstract}

End-to-end motion planning is promising for simplifying complex autonomous driving pipelines. However, challenges such as scene understanding and effective prediction for decision-making continue to present substantial obstacles to its large-scale deployment. In this paper, we present \emph{PIE}, a pioneering framework that integrates advanced perception, reasoning, and intention modeling to dynamically capture interactions between the ego vehicle and surrounding agents. It incorporates a bidirectional Mamba fusion that addresses data compression losses in multimodal fusion of camera and LiDAR inputs, alongside a novel reasoning-enhanced decoder integrating Mamba and Mixture-of-Experts to facilitate scene-compliant anchor selection and optimize adaptive trajectory inference. \emph{PIE} adopts an action-motion interaction module to effectively utilize state predictions of surrounding agents to refine ego planning. The proposed framework is thoroughly validated on the NAVSIM benchmark. \emph{PIE}, without using any ensemble and data augmentation techniques, achieves an 88.9 PDM score and 85.6 EPDM score, surpassing the performance of prior state-of-the-art methods. Comprehensive quantitative and qualitative analyses demonstrate that PIE is capable of reliably generating feasible and high-quality ego trajectories.
%These results demonstrate the potential of integrating advanced reasoning, prediction, and planning methodologies into end-to-end autonomous driving pipelines.

% \textcolor{red}{CoT}

\end{abstract}

%%%%%%%%%%%%%%%%%%%%%%%%%%%%%%%%%%%%%%%               %%%%%%%%%%%%%%%%%%%%%%%%%%%%%%%%%%%%%%%%%%%%%

%% 1. End-to-end autonomous driving: future paradigm for autonomous vehicles

\section{Introduction}

End-to-end motion planning has emerged as a promising paradigm for general robotic systems, including autonomous vehicles (AVs). This data-driven approach has the potential to enable AVs to handle complex and previously unseen scenarios, a capability that becomes increasingly critical as urban environments grow denser and more intricate. By leveraging sensor data directly, end-to-end methods \cite{hu2023planning, Chen2024VADv2EV, chitta2023pami} aim to consolidate the traditionally segmented autonomy pipeline—encompassing perception, prediction, and planning—into a single, cohesive framework.

Despite the promising performance of end-to-end methods, several key challenges remain. First, fusing multimodal data (e.g., image and LiDAR inputs) often leads to compression-induced losses when reducing historical information or fusing features from different sources. Second, while data-driven approaches offer the potential for enhancing environmental understanding, the complexity of real-world driving requires more sophisticated models that are capable of both reasoning and dynamically adjusting their strategies. Third, incorporating the predictions of other traffic participants into the end-to-end planning pipeline often introduces substantial computational overhead. Developing efficient methods to seamlessly integrate these predictions into the planning process remains an open challenge, presenting considerable opportunities for further advancements.

To address these issues, we present \emph{PIE}, an encoder-decoder framework designed to model the interaction between the action of ego vehicle and the motion of nearby agents and to enable more nuanced reasoning about the driving environment. Our approach mitigates data loss and integrates prediction and planning effectively. The contributions of this work are threefold:

\begin{enumerate}
    \item \textbf{Bidirectional Mamba Fusion} We introduce a bidirectional Mamba fusion that effectively improves the multimodal data fusion between camera and LiDAR. A notable improvement of 1.9 PDM score can be achieved by merely employing this fusion approach based on the Transfuser backbone.

    \item \textbf{Reasoning-Enhanced Decoder} To improve scene reasoning in complex driving scenarios, we design an efficient decoder integrating the MoE, harnessing Mamba to enhance trajectory generation.
    
    \item \textbf{Action-Motion Interaction} We propose an action-motion interaction module via a shared cross-attention that directly integrates the velocity predictions of surrounding agents into ego action to model the dynamic interactions between traffic users. 
    
\end{enumerate}

Our approach surpasses the previous state-of-the-art DiffusionDrive \cite{liao2025diffusiondrivetruncateddiffusionmodel} by achieving an 88.9 PDM score and 85.6 EPDM score on the NAVSIM \texttt{navtest} split, demonstrating the superiority and effectiveness of the proposed modules.

\section{Related Work}
Aiming to unify perception and planning within a single framework, end-to-end learning paradigms have attracted increasing attention in the field of autonomous driving. UniAD \cite{hu2023planningorientedautonomousdriving}, as a seminal effort, improves planning performance by integrating multiple perception tasks, thereby demonstrating the potential of end-to-end autonomous driving. VAD\cite{jiang2023vadvectorizedscenerepresentation} further explores compact vectorized scene representations to enhance computational efficiency. Subsequently, a series of studies\cite{chitta2022transfuserimitationtransformerbasedsensor, li2024egostatusneedopenloop,wang2023drivingfuturemultiviewvisual, Weng_2024_CVPR, zheng2024genadgenerativeendtoendautonomous} adopts a single-trajectory planning paradigm to further boost planning quality. VADv2 \cite{Chen2024VADv2EV} is the first to shift toward multi-modal planning by scoring and sampling from a large, fixed vocabulary of anchor trajectories. HydraMDP \cite{li2024hydramdp} refines VADv2’s scoring mechanism by introducing additional supervision from a rule-based scorer. SparseDrive \cite{Sun2024SparseDriveEA} investigates an alternative solution that dispenses with an intermediate BEV representation. DiffusionDrive \cite{liao2025diffusiondrivetruncateddiffusionmodel} employs a truncated diffusion policy for multi-modal trajectory planning. WoTE \cite{li2025endtoenddrivingonlinetrajectory} proposes online trajectory evaluation using a Bird’s-Eye-View (BEV) world model.

Recent efforts to improve computational efficiency in Transformer-based architectures have led to innovations like Mamba-2 \cite{dao2024transformers}, which reduces the quadratic complexity associated with attention computations. Mamba-based architectures have been successfully applied to language modeling \cite{lieber2024jamba, anthony2024blackmamba, he2024densemamba}, time-series forecasting \cite{xu2024integrating, ahamed2024timemachine}, human motion generation \cite{zhang2025motion}, and vision tasks \cite{zhu2024vision, dong2024fusion}. Notably, DRAMA \cite{yuan2024drama} utilized Mamba to fuse image and LiDAR data and use it as a part of the decoder for AV applications. In this work, we propose a bidirectional Mamba fusion approach to address the memory forgetting inherent in Mamba and enhance perception performance.

The rapid advancement of LLMs has also been driven by the Mixture-of-Experts (MoE). Scaling model capacity without commensurate computational overhead, MoE architectures have achieved state-of-the-art results not only in NLP \cite{Fedus2022, Lepikhin2020} but also in domains like vision and time-series analysis \cite{Riquelme2021, Li2024a, Pini2023}. We investigate the application of the MoE within the domain of autonomous driving and integrate it into the decoder, leveraging its distinctive capability to enhance planning performance with minimal computational overhead.

Motion prediction is also crucial for autonomous driving. While recent approaches \cite{zhou2023query, shi_mtr++_2023, sun2024controlmtr} have improved prediction accuracy, they often remain computationally expensive and struggle to achieve satisfactory training efficiency. Their integration into fully end-to-end pipelines remains an area requiring further exploration. These gaps, coupled with the need for enhanced inference, reduced complexity, and effective multimodal fusion, motivate our exploration of efficient and effective frameworks. Given these constraints, we propose predicting the simplified motion of agents within the current frame and employing a cross-attention to integrate ego action and agents' motions, thereby achieving both efficient and resilient planning.

\section{Methodology}
\vspace{2em}
\begin{figure*}[htbp]
  \centering
  \includegraphics[width=0.95\linewidth]{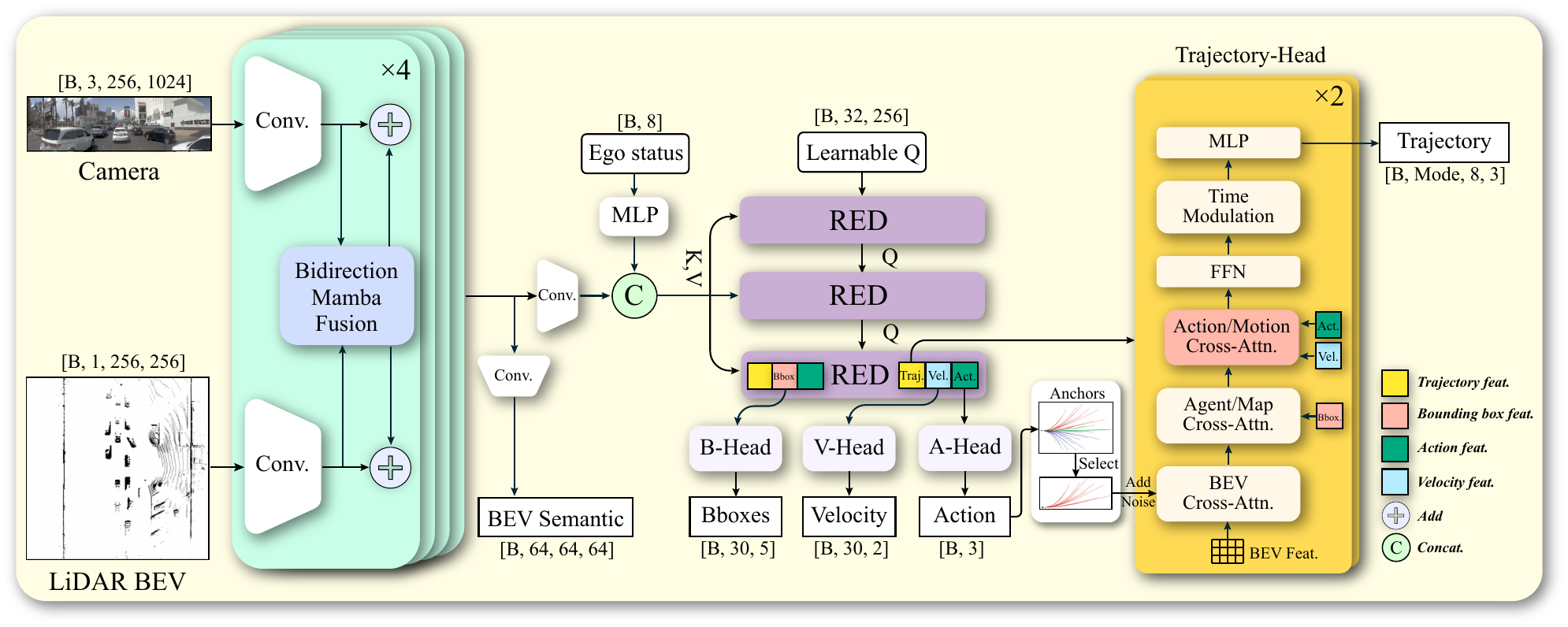}
  \caption{\textbf{Pipeline overview of PIE.} PIE integrates camera and LiDAR BEV images in the feature space, leveraging the Bidirectional Mamba Fusion module for effective fusion in different orders. The fused feature is concatenated with ego status information and passed into the decoder. This decoder, utilizing multiple reasoning-enhanced decoder (RED) layers and specialized heads, outputs a trajectory for autonomous vehicle navigation. Additionally, it predicts the ego's action and agents' bounding boxes and velocities, which are used for auxiliary loss to enhance training and overall model performance.}
  \label{fig:backbone}
\end{figure*}

The overall pipeline of PIE is illustrated in Figure \ref{fig:backbone}. PIE introduces three key modules: Bidirectional Mamba Fusion, Reasoning-Enhanced Decoder (RED), and Intention-Motion Prediction. Initially, multi-view camera images and LiDAR BEV inputs are fused through the bidirectional Mamba fusion module. The fused features, along with ego-vehicle status, are then processed by the reasoning-enhanced decoder, which integrates two core mechanisms: (1) using sequence modeling of Mamba-2 to refine trajectory features; (2) the MoE, which dynamically routes features to specialized expert networks for managing complex driving scenarios. The action-motion interaction module incorporates the state information (bounding box and velocity) of surrounding agents to improve planning accuracy and reliability. The last reason-enhanced decoder layer outputs the ego action and agents' state feature for further action-motion cross attention in the subsequent trajectory head.

\subsection{Bidirectional Mamba Fusion}
According to Mamba-2\cite{dao2024transformers}, for input sequence $x = (x_0, x_1, \dots) \in \mathbb{R}^L$, the process can be written by:

\begin{equation}
    \begin{array}{rcl}
        y_t &=& \sum_{s=0}^t C_t^\top A_{t:s}^\times B_s x_s, \\
    \end{array}   
    \label{eq:cab}
\end{equation}
where $ A_{t:s}^\times$ denotes $A_t \times \cdots \times A_{s+1}, s \leq t$, and $A \in \mathbb{R}^{N \times N} \text{ is parameter matrix and } B \in \mathbb{R}^{N}, C \in \mathbb{R}^{N} $ are parameter vectors and Eq. (\ref{eq:cab}) can be reformulated as:

\begin{equation}
    \begin{array}{rcl}
        y &=& \text{SSM}(A, B, C)(x) = Mx, \\% M_{ji} &\coloneqq& C_j^\top A_j \cdots A_{i+1} B_i
    \end{array}   
    \label{eq:SSM}
\end{equation}
and the matrix $M$ is defined as follows:
\begin{equation}
    \begin{array}{rcl}
       M_{ji} &=& C_j^\top A_j \cdots A_{i+1} B_i
    \end{array}   
\label{eq:m}
\end{equation}

\begin{figure*}[htbp]
  \centering
  \includegraphics[width=0.85\linewidth]{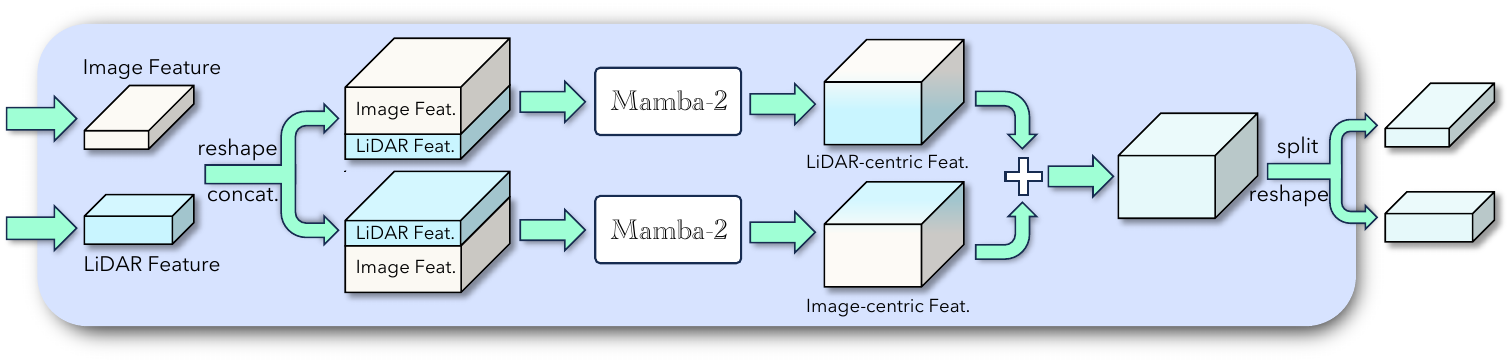}
  \caption{ \textbf{Bidirectional Mamba Fusion.} In this approach, image features and LiDAR features are first reshaped and then concatenated in two different orders. These concatenated features are separately processed by two Mamba-2 modules to generate LiDAR-centric features and Image-centric features, respectively. The outputs of these modules are then combined through an addition operation. Finally, the fused features are split and reshaped back to their original dimensions.}
  \label{fig:fusion}
  \vspace{-1em}
\end{figure*}

Mamba-2 exhibits remarkable proficiency in processing long sequences. Nonetheless, as delineated in Eq. (\ref{eq:cab}), (\ref{eq:SSM}), and (\ref{eq:m}), Mamba-2 leverages the matrix $M$ to compress and retain as much historical data as possible, some loss of information in earlier features is still inevitable due to the limitations inherent in the compression process. To address this limitation, we propose a novel bidirectional fusion method, as illustrated in Figure \ref{fig:fusion}, to effectively integrate image and LiDAR modalities. This method comprises two key components as follows:

% \begin{equation}
%     \begin{array}{rcl}
%         y_t &=& \sum_{s=0}^t C_t^\top A_{t:s}^\times B_s x_s, \\
%     \end{array}   
% \end{equation}

\subsubsection{LiDAR-centric Fusion} Image features are concatenated before LiDAR features along the feature dimension, resulting in a concatenated feature tensor. When processed through Mamba-2, according to its sequential processing mechanism as Eq. (\ref{eq:cab}), the latter part of the sequence (LiDAR features) inherently captures contextual information from the preceding image features. Due to the natural attenuation of earlier information in the feature sequence, this configuration emphasizes LiDAR features while retaining relevant image context.
\subsubsection{Image-centric Fusion} The concatenation order is reversed, with LiDAR features preceding image features. This allows the image features to incorporate contextual information from the LiDAR features while preserving a stronger emphasis on visual information.

The final fused representation is derived through the element-wise addition of these complementary features, resulting in a balanced integration that leverages the strengths of both modalities.

\subsection{Reasoning-enhanced Decoder}

\begin{figure}[htbp]
  \centering
  \includegraphics[width=\linewidth]{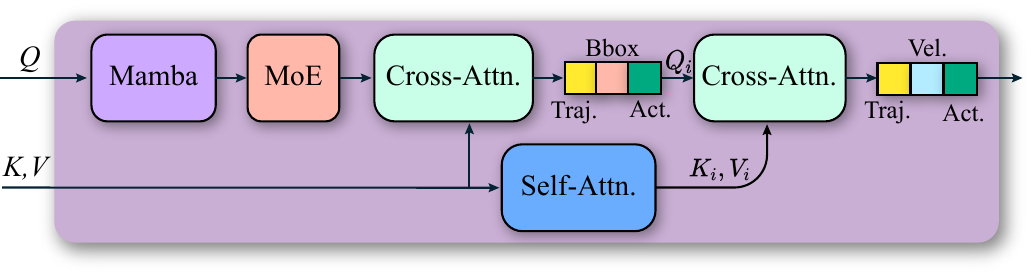}
  
  \caption{\textbf{Reasoning-Enhanced Decoder (RED).} The learnable Q is processed through the Mamba module and MoE, followed by cross-attention with features from the encoder to generate an intermediate feature, part of which is used for bounding box generation. This intermediate feature undergoes further cross-attention with the feature derived through self-attention on the encoder features. The final outputs are the  trajectory, velocity, and action features.}
  \label{fig:RED}
  \vspace{-1.5em}
\end{figure}

The trajectory planning process shown in Figure \ref{fig:RED} utilizes the Mamba-2 architecture combined with a cross-attention mechanism to achieve iterative refinement. Through the first cross-attention, the model generates the trajectory and bounding box feature, and then the second cross-attention layer integrates the output of the first cross-attention layer with environmental and traffic information from the self-attention. We discuss this in more detail in Section \ref{sec:VIM}.

\textbf{Mixture of Experts}
% In real-world traffic scenarios, the complexity and variability of events—such as overtaking manoeuvers and dynamically adjusting following distances—demand flexible and specialized decision-making strategies. 
To equip the planner with the capability to handle complex traffic situations efficiently without incurring excessive computation, we incorporate a Mixture of Experts (MoE) layer immediately following the output layer of Mamba-2. Specifically, the MoE layer comprises two components: a set of experts $E_1, E_2, \dots, E_n$ and a gating network $G$. The MoE layer maintains consistency in the input and output dimensions, serving to refine and enhance the results produced by Mamba-2. For input $x$, the final output $y$ of MoE is denoted as follows:

\begin{equation}
    \begin{array}{rcl}
        y=\sum_{i=0}^{n} G(x)_iE_i(x)
    \end{array}
\end{equation}

Our model uses \textit{Top-K Gating} to select the first $k$ optimal results and integrate results with \textit{Softmax} function. 

\begin{equation}
    \begin{array}{rcl}
        G(x)=Softmax(TopK(x \cdot W_g,k))
    \end{array}
\end{equation}

\begin{equation}
    \begin{array}{rcl}
        TopK(v,k)_i=
        \begin{cases}
        v_i& \text{if $v_i$ is in the top $k$ elements of $v$}\\
        -\infty& \text{otherwise}
        \end{cases}
    \end{array}
\end{equation}
To mitigate overfitting, the gating network normalizes the top $k$ outputs and applies random dropout to the secondary outputs $G_i$. To ensure balanced utilization of the experts, the gating network redistributes any weight exceeding the maximum capacity to the next level of \textit{Top-K} results. In our experimental setting, the number of experts is fixed to 3, and the hidden dimension of the MoE module is set to 768.

\textbf{Anchor Selection}
Planning anchors are vital in generating trajectories that adhere to driving commands and comply with scene constraints. We propose an innovative anchor selection mechanism designed to dynamically identify anchors aligned with the driving command. We cluster three distinct anchor types: turning left, going straight, and turning right, each comprising 20 trajectories. The anchor selection leverages the output of the action head to select the appropriate anchor type.

\subsection{Action-Motion Interaction}

\begin{figure}[htbp]
  \centering
  \vspace{-0.5em}
  \includegraphics[width=\linewidth]{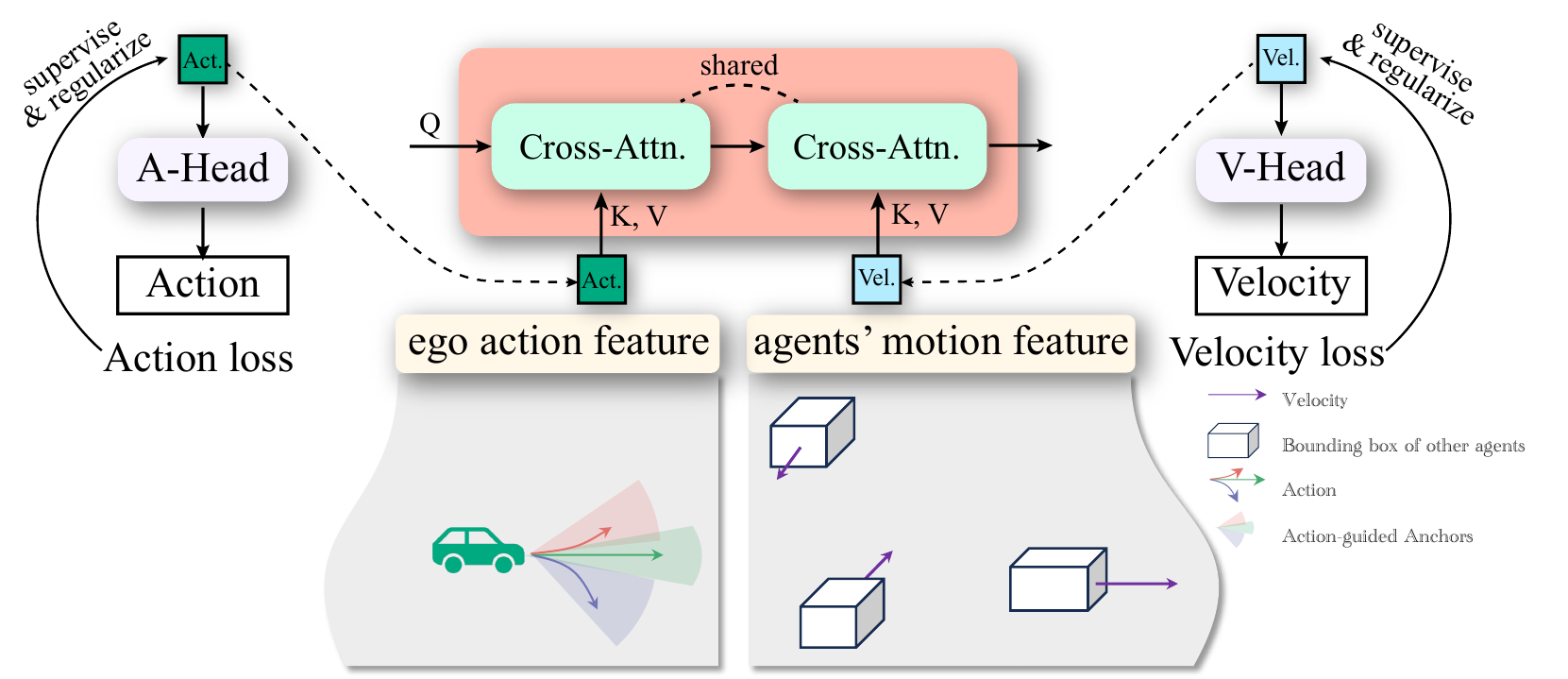}
  \vspace{-1em}
  \caption{\textbf{Action-Motion Cross-Attention.} The ego action feature (Act.) and the agents' motion features (Vel.) are sequentially input into a single shared cross-attention. Each feature is passed to its dedicated decoder for supervision, thereby regularizing its respective latent representation.}
  \label{fig:ActMotion}
\end{figure}

To effectively model the high-level interactions between the ego vehicle and nearby agents without incurring intensive computation, we propose a shared cross-attention designed to enhance planning performance by capturing the dynamic interplay between ego actions and the motions of surrounding agents. As depicted in Figure \ref{fig:ActMotion}, the ego action feature is initially incorporated as key and value within the cross-attention framework, followed by the integration of velocity features from surrounding agents, which are similarly utilized as key and value inputs to the same mechanism.

\subsubsection{Rethink Trajectory Prediction for Planning} Modern prediction models attempt to predict precise trajectories of all agents in the environment. However, such detailed prediction is often unnecessary for real-world expert drivers. Instead, human drivers typically rely on real-time positional and velocity information of nearby agents to make split-second decisions. 

In addition, predicting future states over extended time horizons inherently introduces increased uncertainty, particularly in autonomous driving scenarios. Empirical studies have shown that the variance and multimodality of predicted vehicle trajectories escalate significantly as the prediction horizon lengthens\cite{shao2024uncertainty}. Moreover, scenarios characterized by sparse or long-tail data further exacerbate predictive uncertainty, compelling planning algorithms to adopt overly cautious strategies to maintain safety, thereby compromising efficiency\cite{zhou2022long}. Consequently, these compounded uncertainties inevitably degrade confidence in long-range predictions, directly impairing the accuracy and reliability of the associated planning processes. Given these research findings and drawing inspiration from human decision-making, in addition to bounding box prediction, we also propose to introduce real-time velocity estimations to enhance planning efficacy.

\subsubsection{Velocity Prediction}
\label{sec:VIM}To maintain efficiency, we restrict our predictions to the instantaneous velocities of surrounding agents at the current time step. 
The learnable query processed by the Mamba and MoE module, computes as query vectors $Q$ for cross-attention with BEV features encoded as keys $K$ and values $V$, as shown in Figure~\ref{fig:RED}. Subsequently, the BEV features undergo a self-attention process to extract more velocity information features from the environment, represented as $K_i$ and $V_i$. The intermediate features after the first cross-attention, which are crucial for bounding box generation, serve as query $Q_i$ in the second cross-attention module. The second cross-attention generates the final features, which are then processed by the trajectory and velocity heads in the last RED layer to produce the final trajectory and velocity predictions. The bounding box latent feature after the second cross-attention will be used as velocity feature, hence, these two latent representations will be internally regularized during training, resulting in the velocity feature incorporating the bounding box information. In this context, we designate the velocity feature input into the action-motion cross-attention as the agents' motion feature.

To adapt the model for better velocity prediction, we incorporate the velocity prediction loss within the auxiliary loss framework. This loss component is calculated as the L1 loss between the predicted velocities of agents $V^{p}$ and their corresponding ground truth values $V^{gt}$.
\begin{equation}
     \begin{array}{rcl}
         \mathcal{L}_{Vel} = \mathcal{L}_1(V^{p},V^{gt})\\
     \end{array} 
     \label{eq:loss}
\end{equation}

\subsubsection{Action Re-extraction}
We use the driving command as supervision to re-extract the ego action. The action head, referred to as \textbf{A-Head} in Figure \ref{fig:backbone}, is specifically designed to generate general action intentions, including directives such as left, straight, or right. Beyond its role as an input to the action-motion cross-attention, this action additionally functions as a signal to select scenario-specific anchors tailored to the context. The action is supervised using cross-entropy loss and the loss $\mathcal{L}_{Act}$ is defined as follow:
$$\mathcal{L}_{Act} = \operatorname{CrossEntropy}(Action, DrivingCmd[:3])$$
The reasons for not directly using the driving command are twofold: Firstly, we aim to constrain the action output to a set of three directives, namely, left, straight, and right, explicitly excluding the \textbf{unknown} command, thereby enabling scenario-specific anchor selection. Secondly, we intend for this supervision to enhance the learning of action features, resulting in this latent representation being well leveraged to facilitate the action-motion interaction.

% We empirically observe that using a shared-weight cross-attention module for both ego intention and surrounding agents’ velocity information significantly improves performance compared to using separate modules. It also enables a lightweight, step-wise reasoning process where the model first incorporates the ego's intended action and then refines the representation by attending to dynamic interactions.

\subsection{Loss}
We follow the loss design $\mathcal{L}_{DD}$ as implemented in \cite{liao2025diffusiondrivetruncateddiffusionmodel} and introduce two auxiliary loss components, velocity loss $\mathcal{L}_{Vel}$ and action loss $\mathcal{L}_{Act}$. The overall loss function for our method can be formulated as follows:
$$\mathcal{L} = \mathcal{L}_{DD} + \lambda_{v}\mathcal{L}_{Vel} + \lambda_{a}\mathcal{L}_{Act},$$ where $\lambda_{*}$ represents the weighting coefficients assigned to each respective loss term.

\section{Experiments}

\subsection{Experimental Setup}
\subsubsection{\textbf{Dataset}} Our study utilizes the NAVSIM dataset \cite{dauner2024navsim}, which is based on OpenScene\cite{contributors2023openscene}, and serves as a streamlined version of the nuPlan dataset \cite{caesar2021nuplan}.
This dataset includes extensive driving logs totaling 120 hours and the primary data inputs for the agent comprise high-resolution images (1920$\times$1080 pixels) from multiple viewpoints, coupled with consolidated LiDAR point cloud data derived from five distinct sensors. The input data ensemble includes the current frame along with, optionally, the three preceding frames, spanning a total duration of 1.5 seconds at a 2Hz sampling rate.

\subsubsection{\textbf{Metrics}} The NAVSIM employs a specialized metric, the Predictive Driver Model score (PDM score), to assess the overall performance of end-to-end motion planners. The formula for PDM score is structured as follows:

\begin{equation}
\begin{array}{c}
\text{PDMS} = \left( \frac{5 \times \text{EP} + 5 \times \text{TTC} + 2 \times \text{C}}{12} \right) \times \text{NC} \times \text{DAC,}
\end{array}
\end{equation}
where Ego Progress (EP) quantifies the vehicle's progression along its intended route, focusing on advancement efficiency. Time-to-collision (TTC) assesses safety margins by measuring the temporal distance to potential collision points with other vehicles, thereby evaluating collision risk. Additionally, comfort (C) examines the trajectory’s acceleration and jerk relative to established comfort thresholds, assessing the ride quality. Lastly, NC represents no collision with traffic participants, and DAC denotes no infraction regarding Drivable Area Compliance. This scoring system assigns a PDM score value of zero in instances of collisions or non-compliance with drivable area regulations. 
We also evaluate our performance with the extended PDM score (EPDMS):

\begin{equation}
\begin{array}{c}
\text{EPDMS} = \left( \frac{5 \times \text{EP} + 5 \times \text{TTC} + 2 \times \text{HC} + 2 \times \text{LK} + 2 \times \text{EC}}{16} \right) \vspace{0.4em} \\  \hspace{4em} \times \text{NC} \times \text{DAC} \times \text{DDC} \times \text{TLC.}
\end{array}
\end{equation}
The newly introduced subscores are History Comfort (HC), Lane Keeping (LK), Extended Comfort (EC), Driving Direction Compliance (DDC), and Traffic Light Compliance (TLC), each designed to address specific aspects of driving performance evaluation.
\begin{figure*}[htbp]
  \centering
  \includegraphics[width=0.9\linewidth]{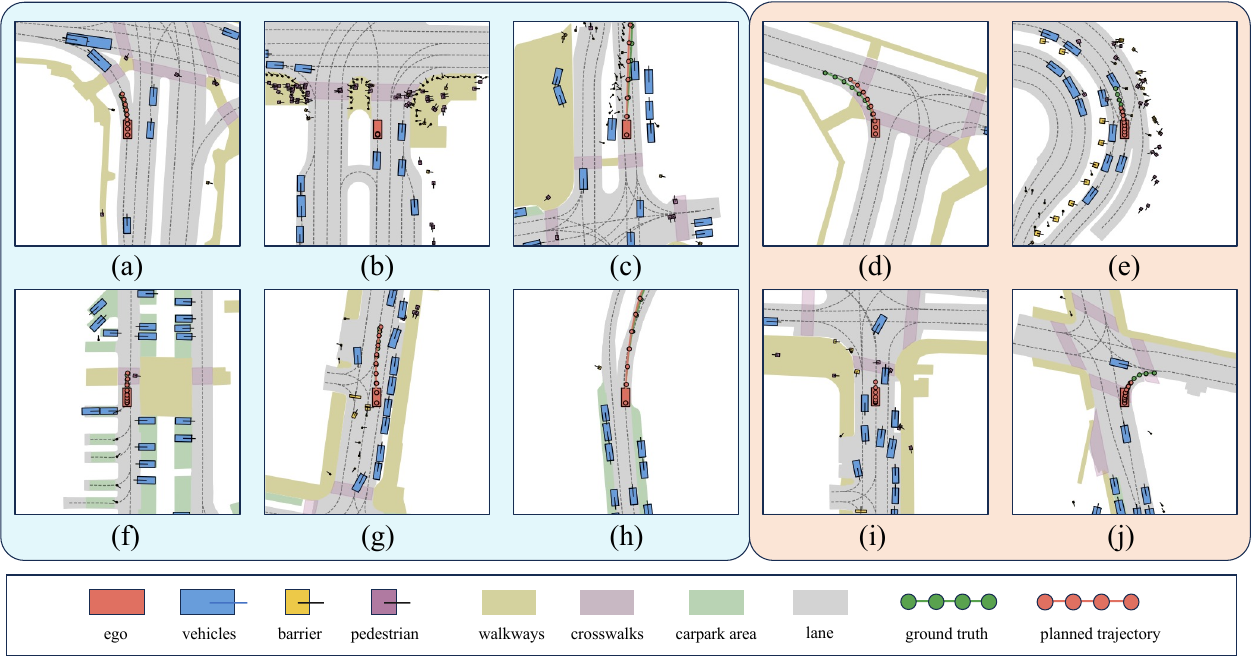}
  \caption{Visualizations of the planning results of PIE across various scenarios: (a) Following a vehicle through a turn (b) Waiting at a red light (c) Lane change for obstacle avoidance (d) Turn left at a T-junction (e) Yielding to pedestrians (f) Crossing a crosswalk with pedestrian (g) Lane change at a T-junction (h) Turning and changing lanes at a fork in the road (i) Slow driving and stopping (j) Yielding right of way to oncoming vehicles when turning right.}

  \label{fig:q_r}
\end{figure*}

\begin{figure}[htbp]
  \centering

  \includegraphics[width=0.95\linewidth]{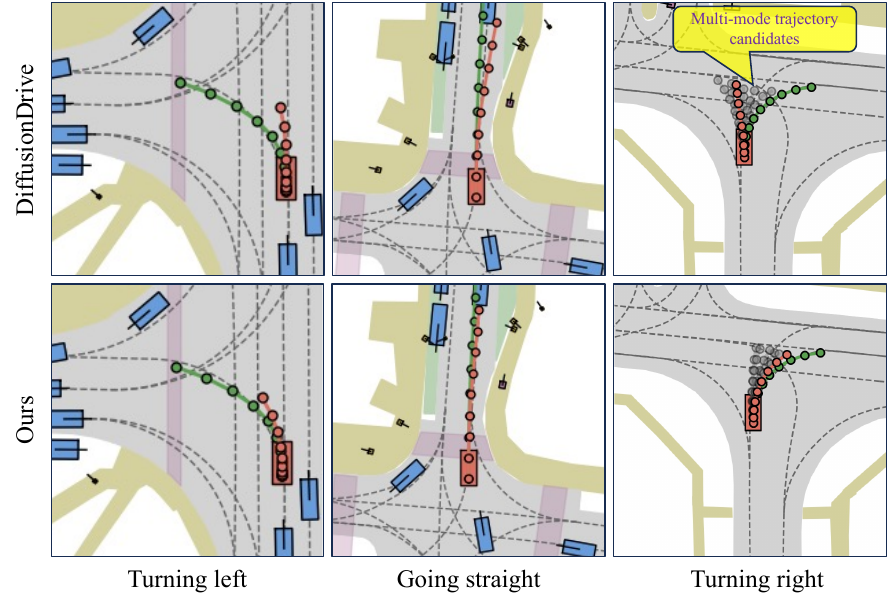}
  \vspace{-1em}
  \caption{\textbf{Qualitative comparison of DiffusionDrive and PIE.} With the anchor selection, our method generates a refined set of scene-compliant anchors, which effectively guide trajectory generation and facilitate more plausible and feasible waypoints.}
  \label{fig:ModeCompare}
  \vspace{-1em}
\end{figure}

\subsubsection{\textbf{Implementation Details}} Our models are trained using the NAVSIM \texttt{navtrain} split and evaluated on the NAVSIM \texttt{navtest} split. Training and testing are conducted on an NVIDIA RTX 3090 Ti GPU with batch size 32, and the training epoch is 100. We use the AdamW optimizer, setting the learning rate and weight decay at 2e-4 and 0.01, respectively. The visual inputs are derived from camera images captured at the front-left and front-right positions, which are center-cropped and subsequently merged with the front-view image, forming a unified input dimension of 256$\times$1024 pixels. The LiDAR-based BEV image is constructed by projecting LiDAR points onto a BEV plane. We only utilize the contemporaneous camera and LiDAR imagery as inputs, without historical frames and data augmentation. Additionally, the model integrates real-time data on the ego vehicle's dynamics, such as velocity and acceleration, along with navigational commands like turning, lane changing, and following. The final output is an 8-point trajectory over a 4-second horizon, sampled at 2 Hz, where each waypoint includes coordinates for x, y, and orientation.

\subsection{Qualitative Result}

%%%%%%%%%%%%%%%%%%
The proposed planner’s performance is illustrated through ten diverse traffic scenarios in Figure \ref{fig:q_r}, with several examples (highlighted in orange block) surpassing human driving behavior.

In subfigure (a), the planner ensures safer left turns by maintaining a trajectory closer to the turn lane’s centerline, enhancing safety over human drivers. Subfigure (b) demonstrates the planner’s ability to perceive complex intersections, halting for pedestrians in compliance with traffic laws. Subfigure (c) addresses a challenging scenario with barricades and pedestrians. The planner identifies obstacles and executes a lane change to maintain safety and speed.

In subfigure (f), the planner correctly identifies that crossing pedestrians do not compromise safety, maintaining velocity without unnecessary deceleration. Subfigures (g) and (h) showcase robust planning capabilities, with (g) dynamically navigating a T-junction to avoid congestion and narrow roads and (h) executing smooth lane changes at road forks.

Subfigures (d), (e), (i), and (j) further showcase superior decision-making. Subfigure (d) achieves better lane positioning, maintaining a safer distance from the lane edge. In subfigure (e), the planner adopts a cautious approach, yielding to pedestrians in a U-shaped road segment. Subfigure (i) optimizes parking near crosswalks, and subfigure (j) ensures safer right turns by slowing down.

These examples underscore the planner’s effectiveness in navigating complex urban traffic, demonstrating advanced perception, reasoning, and adaptability in scenarios that challenge even skilled human drivers.

In Figure \ref{fig:ModeCompare}, we conduct a qualitative comparison between DiffusionDrive and our proposed method. We evaluate three distinct driving scenarios: left turns, going straight, and right turns. Compared with DiffusionDrive, our planning results exhibit better feasibility and adherence to navigation information, ascribing to our innovative anchor selection mechanism. Especially shown in the right-turn scenario, our generated multimodal trajectory candidates demonstrate superior compliance with driving commands while effectively mitigating mode collapse.

\begin{table*}
\centering
\caption{Performance Comparison on Navtest Split}
\label{tab:Comparsion_v1} 
\begin{tabular}{c|ccc|ccccccc}  % 确保列的定义与表头和内容相匹配
\hline

Method & Inputs & Img.Backbone & NC $\uparrow$ & DAC $\uparrow$ & EP $\uparrow$ & TTC $\uparrow$ & C $\uparrow$ & PDMS $\uparrow$ \\
\hline
UniAD\cite{hu2023planningorientedautonomousdriving}  & Camera & ResNet-34 & 97.8 & 91.9 & 78.8 & 92.9 & \textbf{100} & 83.4 \\
PARA-Drive\cite{Weng_2024_CVPR}  & Camera & ResNet-34 & 97.9 & 92.4 & 79.3 & 93.0 & 99.8 & 84.0 \\
LTF\cite{chitta2022transfuserimitationtransformerbasedsensor}  & Camera & ResNet-34 & 97.4 & 92.8 & 79.0 & 92.4 & \textbf{100} & 83.8 \\
Transfuser\cite{chitta2022transfuserimitationtransformerbasedsensor}  & Camera \& LiDAR & ResNet-34 & 97.7 & 92.8 & 79.2 & 92.8 & \textbf{100} & 84.0 \\
DRAMA\cite{yuan2024drama}  & Camera \& LiDAR & ResNet-34  & 98.0 & 93.1 & 80.1 & 94.8 & \textbf{100} & 85.5 \\
Hydra-MDP\cite{li2024hydramdpendtoendmultimodalplanning} & Camera \& LiDAR & ResNet-34  & \underline{98.3} & 96.0 & \underline{}78.7 & 94.6 & \textbf{100} & 86.5 \\
DiffusionDrive\cite{liao2025diffusiondrivetruncateddiffusionmodel} & Camera \& LiDAR & ResNet-34  & 98.2 & 96.2 & \underline{82.2} & 94.7 & \textbf{100} & 88.1 \\
WoTE\cite{li2025endtoenddrivingonlinetrajectory} & Camera \& LiDAR & ResNet-34  & \textbf{98.5} & \underline{96.8} & 81.9 & \underline{94.9} & \underline{99.9} & \underline{88.3} \\

PIE (Ours) & Camera \& LiDAR & ResNet-34 & \underline{98.3} & \textbf{96.9} & \textbf{83.0} & \textbf{95.0} & \textbf{100} & \textbf{88.9} \\

\hline
\end{tabular}
\end{table*}

\begin{table*}
\centering
\caption{Performance Comparison on Navtest Split with Extended PDMS}
\label{tab:EPMS} 
\begin{tabular}{c|cccccc|c}  % 确保列的定义与表头和内容相匹配
\hline

Method & NC $\uparrow$ & DAC $\uparrow$ & EP $\uparrow$ & TTC $\uparrow$ & DDC $\uparrow$ & LK $\uparrow$& EPDMS $\uparrow$ \\
\hline
Transfuser\cite{chitta2022transfuserimitationtransformerbasedsensor}  & 97.7 & 92.8 & 79.2 & 92.8 & 98.3 & 67.6 & 77.8 \\
VADv2\cite{Chen2024VADv2EV} & 97.3 & 91.7 & 77.6 & 92.7 & 98.2 & 66.0 & 76.6 \\
Hydra-MDP \cite{li2024hydramdpendtoendmultimodalplanning} & 97.5 & 96.3 & 80.1 & 93.0 & 98.3 & 65.5 & 79.8 \\

Hydra-MDP++ \cite{li2025hydramdpadvancingendtoenddriving} & 97.9 & 96.5 & 79.2 & 93.4 & 98.9 & 67.2 & 80.6 \\

ARTEMIS \cite{feng2025artemisautoregressiveendtoendtrajectory} & \textbf{98.3} & 95.1 & 81.5 & \underline{97.4} & 98.6 & 96.5 & 83.1 \\

DiffusionDrive\cite{liao2025diffusiondrivetruncateddiffusionmodel} & \underline{98.2} & 96.2 & \textbf{87.6} & 97.3 & 98.6 & 97.0 & 84.0 \\

GaussianFusion \cite{liu2025gaussianfusiongaussianbasedmultisensorfusion} & \textbf{98.3} & \textbf{97.3} & \underline{87.5} & \underline{97.4} & \underline{99.0} & \textbf{97.4} & \underline{85.0} \\

PIE (Ours) & \textbf{98.3} & \underline{96.9} & \textbf{87.6} & \textbf{97.6} & \textbf{99.5} & \underline{97.2} & \textbf{85.6} \\

\hline
\end{tabular}
\begin{tablenotes}

\item \noindent Under the rapid-iteration release of the NAVSIM benchmark, the subscore naming has been somewhat ambiguous. For clarity of comparison, we additionally report the following subscores for our method: TL = 99.8, HC = 98.3, and EC = 88.4. 
\end{tablenotes}
\end{table*}

\subsection{Quantitative Results}

\begin{table*}
    
\centering
\caption{Effectiveness Evaluation of the Proposed Modules}
\label{tab:effectiveness_evaluation}
\vspace{-3pt}
% \resizebox{\linewidth}{!}{
\begin{tabular}{c|c|c|cccccc}
\hline
& Method & Para. & NC $\uparrow$ & DAC $\uparrow$ & EP $\uparrow$ & TTC $\uparrow$ & C $\uparrow$ & PDMS $\uparrow$ \\
\hline
\multirow{2}{*}{Baseline} & Transfuser(T) & 56.0M & 97.7 & 92.8 & 79.2 & 92.8 & \textbf{100} & 84.0 \\
& DiffusionDrive(DD) & 60.0M & 98.2 & 96.2 & 82.2 & 94.7 & \textbf{100} & 88.1 \\
\hline
\multirow{7}{*}{Ours}

%& D & 0.975 & 0.923 & 0.787 & 0.941 & 1 & 0.843 \\

& T + RED & 58.0M & 97.4 & 93.6 & 79.5 & 92.6 & \textbf{100} & 84.4 \\
& T + V.P. & 56.7M & 97.4 & 93.8 & 79.2 & 93.0 & \textbf{100} & 84.5 \\
& T + BDM & 57.7M & 98.0 & 94.3 & 80.6 & 93.6 & \textbf{100} & 85.9 \\
& DD + BDM & 61.9M & 98.2 & 96.6 & 82.7 & 94.4 & \textbf{100} & 88.3 \\
& DD + BDM + RED & 63.9M & 98.2 & 96.6 & 82.7 & 94.7 & \textbf{100} & 88.5 \\
& DD + AMI & 64.8M & \textbf{98.3} & 96.8 & \textbf{83.0} & 94.7 & \textbf{100} & 88.7 \\
%& DRAMA-II & \textbf{0.983} & \textbf{0.948} & \textbf{0.810} & 0.942 & 1 & \textbf{0.866} \\
& PIE & 68.5M & \textbf{98.3} & \textbf{96.9} & \textbf{83.0} & \textbf{95.0} & \textbf{100} & \textbf{88.9} \\

\hline
\end{tabular}
\begin{tablenotes}

\item \noindent RED: Reasoning-Enhanced Decoder; V.P.: Velocity Prediction (only velocity prediction, no action-motion interaction); BDM: Bidirectional Mamba Fusion Module, using two stacked Mamba-2 layers for each data fusion branch; AMI: Action-Motion Interaction.
\end{tablenotes}
\end{table*}

In this section, we provide a quantitative comparison between our method and existing approaches, followed by an analysis of the efficacy of the proposed modules and ablation studies on different experimental settings.

To enable a fair comparison, we selected other advanced end-to-end driving models that use the same image backbone. On the NAVSIM navtest split, PIE achieves state-of-the-art performance across both PDMS and EPDMS metrics. As shown in Table \ref{tab:Comparsion_v1}, compared to DiffusionDrive, PIE achieves better performance on all metrics, especially on DAC (+0.7) and EP (+0.8). When using the extended PDM score (presented in Table \ref{tab:EPMS}), our method achieves the highest DDC subscore (99.5), demonstrating the effectiveness of the anchor selection mechanism.

We evaluate the effectiveness of our proposed modules in Table~\ref{tab:effectiveness_evaluation}. Compared to the baseline Transfuser (T) with a PDM score of 84.0 and DiffusionDrive (DD) with a PDM score of 88.1, the integration of individual modules such as RED, V.P., BDM, and AMI into PIE yields quantitative enhancements. As illustrated in the table, the addition of the BDM fusion module to Transfuser significantly boosts the PDM score to 85.9. This substantial performance gain indicates the module effectively fuses LiDAR and image modalities through bidirectional fusion, preserving more modality information and strengthening the model’s environmental perception. When all modules are combined in PIE, the model achieves the highest PDM score of 88.9, along with improvements in all key metrics (NC, DAC, EP, TTC).

% \subsection{Ablation Study}

\begin{table*}
\caption{Ablation study on Bidirectional Mamba Fusion}
\label{tab:BDM}
\centering

\begin{tabular}{c|c|c|ccccc|c}
\hline
Method & Mamba layer(s) for single fusion & Para. & NC $\uparrow$ & DAC $\uparrow$ & EP $\uparrow$ & TTC $\uparrow$ & C $\uparrow$ & PDMS $\uparrow$ \\
\hline

%\multirow{8}{*}{Ours} Layer in Each Block

%& D & 0.975 & 0.923 & 0.787 & 0.941 & 1 & 0.843 \\

Transfuser(T) & N.A. & 56.0M & 97.7 & 92.8 & 79.2 & 92.8 & \textbf{100} & 84.0 \\
\hline
T + BDM (++) & 1 & 53.1M & 97.3 & 91.8 & 78.1 & 92.7 & \textbf{100} & 83.0  \textcolor{cyan}{-1.0}\\
T + BDM (++) & 2 & 57.7M & 97.5 & 92.5 & 78.7 & 92.5 & \textbf{100} & 83.6  \textcolor{cyan}{-0.4}\\
T + BDM (+-)& 1 & 53.1M & \textbf{98.2} & 93.8 & 80.0 & \textbf{93.7} & \textbf{100} & 85.4 \textcolor{red}{+1.4} \\
T + BDM (+-)& 2 & 57.7M & 98.0 & \textbf{94.3} & \textbf{80.6} & 93.6 & \textbf{100} & \textbf{85.9}   \textcolor{red}{+1.9}\\

%& DRAMA-II & \textbf{0.983} & \textbf{0.948} & \textbf{0.810} & 0.942 & 1 & \textbf{0.866} \\
\hline
\end{tabular} 

\begin{tablenotes}

\item \noindent + : LiDAR-centric fusion;  - : Image-centric fusion. ++: two fusions are both LiDAR-centric; +-: one LiDAR-centric and one Image-centric fusion. Details of BDM design are illustrated in Figure \ref{fig:fusion}.
\end{tablenotes}

\end{table*}

% \begin{table}
% \caption{Ablation study on Bidirectional Mamba Fusion}
% \label{tab:BDM}
% \centering
% \resizebox{\linewidth}{!}{
% \begin{tabular}{c|cc|cccccc}
% \hline
% Method & Para. & L/B & NC $\uparrow$ & DAC $\uparrow$ & EP $\uparrow$ & TTC $\uparrow$ & C $\uparrow$ & PDM Score $\uparrow$ \\
% \hline

% %\multirow{8}{*}{Ours} Layer in Each Block

% %& D & 0.975 & 0.923 & 0.787 & 0.941 & 1 & 0.843 \\

% DRAMA (D)   & 50.4M & 1 & 0.977 & 0.929 & 0.790 & 0.927 & \textbf{1} & 0.840 \\
% \hline
% D + BDM (++) & 53.1M & 1 & 0.973 & 0.918 & 0.781 & 0.927 & \textbf{1} & 0.830 \\
% D +2BDM (++) & 57.7M & 2 & 0.975 & 0.925 & 0.787 & 0.925 & \textbf{1} & 0.836 \\
% D + BDM (+-)& 53.1M & 1 & \textbf{0.982} & 0.938 & 0.800 & \textbf{0.937} & \textbf{1} & 0.854 \\
% D +2BDM (+-)& 57.7M & 2 & 0.980 & \textbf{0.943} & \textbf{0.806} & 0.936 & \textbf{1} & \textbf{0.859} \\

% %& DRAMA-II & \textbf{0.983} & \textbf{0.948} & \textbf{0.810} & 0.942 & 1 & \textbf{0.866} \\
% \hline
% \end{tabular} 
% }
% \begin{tablenotes}

% \item \noindent + : LiDAR-centric fusion;  - : Image-centric fusion. Details are illustrated in Figure \ref{fig:fusion}.
% \end{tablenotes}

% \end{table}

% \subsection{Ablation Study}

% \subsection{Ablation Study}

\begin{table}
\caption{Ablation study on whether to use shared cross-attention}
\label{tab:Shared}
\centering
\resizebox{\linewidth}{!}{
\begin{tabular}{c|ccccc|c}
\hline
Cross-Attn. & NC $\uparrow$ & DAC $\uparrow$ & EP $\uparrow$ & TTC $\uparrow$ & C $\uparrow$ & PDMS $\uparrow$ \\ 
\hline

Unshared & 98.28 & 96.67 & 82.88 & 94.48 & \textbf{100} & 88.50\\
Shared & \textbf{98.32} & \textbf{96.92} & \textbf{82.99} & \textbf{95.00} & \textbf{100} & \textbf{88.89}  \\

\hline
\end{tabular} 
}
% \begin{tablenotes}

% \item \noindent + : LiDAR-centric fusion;  - : Image-centric fusion. ++: two fusions are both LiDAR-centric; +-: one LiDAR-centric and one Image-centric fusion. Details of BDM design are illustrated in Figure \ref{fig:fusion}.
% \end{tablenotes}

\end{table}

\begin{table}
\caption{Ablation study on number of experts in MoE }
\label{tab:MoE}
\centering

\begin{tabular}{c|c|c|c|c}
\hline
Num of experts & Dim. & Para. & PDMS $\uparrow$  &FPS $\uparrow$ \\  
\hline

2 & 512 & 66.5M  & 88.75 & \textbf{39}  \\
3 & 768 & 68.5M  & \textbf{88.89} & 37\\
4 & 1024 & 71.2M  & 88.78 & 34 \\

\hline
\end{tabular} 

\begin{tablenotes}

\item \noindent The FPS results were obtained using a single NVIDIA RTX 4090 GPU.
\end{tablenotes}
\vspace{-2em
}
\end{table}

To evaluate the effectiveness of our Bidirectional Mamba fusion (BDM) design, we conducted an ablation study, as presented in Table \ref{tab:BDM}, in which we concatenated the image and LiDAR tensors into two sequences without reversing one of them, thus removing the bidirectional property. Under these conditions, the performance declined across multiple PDM score subscores, indicating that using two LiDAR-centric fusion is detrimental. In contrast, using one LiDAR-centric and one Image-centric fusion resulted in a significant improvement in model performance across all metrics. The variant with two Mamba-2 layers in each fusion branch of the BDM achieved the highest score of 85.9. 

Additionally, we conducted an ablation study in Table \ref{tab:Shared} to evaluate the impact of employing shared cross-attention during the action-motion interaction process. The adoption of a single shared cross-attention mechanism yielded a notable increase in PDM score, with an improvement of 0.39.

We evaluated the impact of changing the number of experts in Table\ref{tab:MoE}. The results demonstrate that using three experts with a dimension of 768 achieves an optimal balance between performance and efficiency.

\section{Conclusion}

In this study, we propose PIE, an advanced framework designed to enhance perception and dynamic interaction for end-to-end autonomous driving. Its reasoning-enhanced decoder facilitates the selection of scene-compliant anchors and enhances the generation of feasible trajectories. The bidirectional Mamba fusion exhibits its efficacy through a significant performance enhancement (+1.9 PDM score), achieved by integrating it into the baseline Transfuser model. The proposed interaction mechanism leverages predictions from other agents to model the intention-level interaction between the ego and nearby agents, enhancing the planning quality. We assess the performance of PIE on the NAVSIM leaderboard, where it achieves a PDM score of 88.9 and an EPDM score of 85.6, outperforming the previous state-of-the-art methods, DiffusionDrive (PDMS 88.1) and GaussianFusion (EPDMS 85.0), respectively. The quantitative and qualitative results demonstrate the effectiveness and robustness of our method.

%%%%%%%%%%%%%%%%%%%%%%%%%%%%%%%%%%%%%%%               %%%%%%%%%%%%%%%%%%%%%%%%%%%%%%%%%%%%%%%

% \section*{Acknowledgments}
% This should be a simple paragraph before the References to thank those individuals and institutions who have supported your work on this article.

%%%%%%%%%%%%%%%%%%%%%%%%%%%%%%%%%%%%%%%               %%%%%%%%%%%%%%%%%%%%%%%%%%%%%%%%%%%%%%%%%%%%%

% {\appendix[Proof of the Zonklar Equations]
% Use $\backslash${\tt{appendix}} if you have a single appendix:
% Do not use $\backslash${\tt{section}} anymore after $\backslash${\tt{appendix}}, only $\backslash${\tt{section*}}.
% If you have multiple appendixes use $\backslash${\tt{appendices}} then use $\backslash${\tt{section}} to start each appendix.
% You must declare a $\backslash${\tt{section}} before using any $\backslash${\tt{subsection}} or using $\backslash${\tt{label}} ($\backslash${\tt{appendices}} by itself
%  starts a section numbered zero.)}

%{\appendices
%\section*{Proof of the First Zonklar Equation}
%Appendix one text goes here.
% You can choose not to have a title for an appendix if you want by leaving the argument blank
%\section*{Proof of the Second Zonklar Equation}
%Appendix two text goes here.}

% \section{References Section}

% argument is your BibTeX string definitions and bibliography database(s)
%\bibliography{IEEEabrv,../bib/paper}
%
% \section{Simple References}
% You can manually copy in the resultant .bbl file and set second argument of $\backslash${\tt{begin}} to the number of references
%  (used to reserve space for the reference number labels box).

\bibliographystyle{IEEEtran}
\bibliography{root}
\end{document}